\documentclass{article}
\usepackage{amsmath}
\usepackage{amssymb}
\usepackage{bbm}
\usepackage{verbatim}
\usepackage{authblk}
\usepackage{amsthm}
\usepackage[style=nature, backend=bibtex, sorting=none]{biblatex}
\numberwithin{equation}{section}
\nocite{*}
\begin{document}
\title{Evolution Strategies Converges to Finite Differences}
\author{J. C. Raisbeck}
\author{Matthew Allen}
\author{Ralph Weissleder}
\author{Hyungsoon Im}
\author{Hakho Lee}
\date{December 27, 2019}
\affil{Center for Systems Biology, Massachusetts General Hospital}
\maketitle
\noindent

\textbf{Abstract:  }
Since the debut of Evolution Strategies (ES) as a tool for Reinforcement Learning by Salimans et al. 2017\cite{salimans2017}, there has been interest in determining the exact relationship between the Evolution Strategies gradient and the gradient of a similar class of algorithms, Finite Differences (FD).\cite{zhang2017}\cite{lehman2018} Several investigations into the subject have been performed, investigating the formal motivational differences\cite{lehman2018} between ES and FD, as well as the differences in a standard benchmark problem in Machine Learning, the MNIST classification problem\cite{zhang2017}. This paper proves that while the gradients \textit{are} different, they converge as the dimension of the vector under optimization increases.

\section{Introduction}
Evolutionary computation has been a subject of research in Machine Learning since the middle of the last century, with some methods recently seeing success in the subfield of Reinforcement Learning. One such algorithm is known as Evolution Strategies. Although this name has referred to a larger class of algorithms in the past, this paper uses it exclusively to refer to the specification of the Natural Evolution Strategies framework\cite{wierstra2014} described by Salimans et al.\cite{salimans2017}.

Natural Evolution Strategies (NES) is a class of black-box evolutionary methods for policy optimization which works by controlling parameterized \textit{distributions of policies}, improving by moving parameters of the distribution in accordance with the gradient $\nabla\mathbb{E}_{\theta}[f(\textbf{z})]$, where $\theta$ is the set of parameters of the distribution. Evolution Strategies specifies this framework by optimizing distributions of \textit{parameterized policies}, taking one such parameterized policy as its distribution parameter set $\theta$. ES then optimizes the expectation of the Gaussian distribution with mean $\theta$, and standard deviation $\sigma$.

Because the gradient approximation derived by Salimans et al.\cite{salimans2017} for ES is simple to implement and empirically effective, it has been the subject of much of the research within the NES framework. This formulation is also similar to one version of Finite Differences, which has been a source of some concern. Salimans et al. consider this in the context of the ``curse of dimensionality'', noting that ES might be better at optimizing in high-dimensional space, however they do not investigate to what degree the algorithms are similar.

This paper augments two works which directly investigate these similarities, both published by the UberAI laboratory. Lehman et al.\cite{lehman2018} attempted to differentiate the algorithms by elaborating on the different motivations of the two, and by providing examples. Zhang et al.\cite{zhang2017} considered the differences in the context of the MNIST character classification problem. This paper provides a proof of motivational convergence which explains the similarity noted by Zhang et al., and explains how motivational distinctions noted by Lehman et al. fade in high-dimension optimization problems.

\section{Notation}
Let
\begin{enumerate}
	\item $R$ be a function; $R : \mathbb{R}^{n} \rightarrow \mathbb{R}$,
	\item $\theta$ and $\alpha$ be vectors of dimension $n$; $\theta, \alpha \in \mathbb{R}^{n}$,
	\item $I$ be the identity matrix of dimension $n$,
	\item $X$ be a set of vectors $\alpha_{i}$,
	\item $\lambda = |X|$\footnote{A slightly more familiar equivalent notation might be $\frac{1}{m}\sum_{i=1}^{m}$}, and
	\item $\sigma$ be a positive real number.
\end{enumerate}

\subsection{Gradients}
In order to determine the relationship between Finite Differences and Evolution Strategies, it is useful to examine the core of each algorithm, the gradient approximation. These are described below. In each case, $\alpha \thicksim \mathcal{N}(\theta, \sigma^{2} I)$.

\subsubsection{Finite Differences}
The gradient used in Finite Differences is
\begin{align}
	\nabla R(\theta)\approx	\frac{1}{\lambda}\sum_{\alpha \in X}\frac{(\alpha - \theta)}{||\alpha - \theta||}\frac{R(\alpha) - R(\theta)}{||\alpha - \theta||}.\label{FDgrad}
\end{align}

\subsubsection{Evolution Strategies}
The gradient used in Evolution Strategies\cite{salimans2017} is
\begin{align}
	\nabla \mathbb{E}(R(\pi(\alpha  | \theta)))\approx\frac{1}{\lambda}\sum_{\alpha \in X} (\alpha - \theta) R(\alpha).\label{ESgrad}
\end{align}

\section{The Proof\label{proof}}
Suppose that $||\alpha-\theta|| = ||\mathcal{N}(\vec{0},\sigma^{2} I)||$ is tightly distributed about $\mathbb{E}[||\mathcal{N}(\vec{0}, \sigma^{2} I)||]$. Then,

\begin{align}
	\mathbb{E}[||\mathcal{N}(\vec{0}, \sigma^{2} I)||]^{2}\frac{1}{\lambda}\sum_{\alpha \in X}\frac{(\alpha - \theta)}{||\alpha - \theta||}\frac{R(\alpha) - R(\theta)}{||\alpha - \theta||} 
	\approx \frac{1}{\lambda}\sum_{\alpha \in X}(\alpha - \theta)(R(\alpha) - R(\theta))\label{ESlin}
\end{align}
that is, \ref{ESlin} is linearly dependent upon \ref{ESgrad}.\\
Now, using a similar line of reasoning to that applied in \textsection 3.2 of \textit{Evolution Strategies as a Scalable Alternative to Reinforcement Learning}\cite{salimans2017} consider the difference between this approximation and that of ES:
\begin{align}
	& \frac{1}{\lambda}\sum_{\alpha \in X}(\alpha - \theta)(R(\alpha) - R(\theta)) 
	- \frac{1}{\lambda}\sum_{\alpha \in X} (\alpha - \theta) R(\alpha)\\
	= & \frac{1}{\lambda}\sum_{\alpha \in X}(\alpha - \theta)(- R(\theta))\\
	= & - \frac{1}{\lambda}R(\theta)\sum_{\alpha \in X}(\alpha - \theta)
\end{align}
That is, the difference $\nabla FD - \nabla ES$ is a $\mathcal{N}(\vec{0}, R(\theta)^{2}(\frac{\sigma}{\lambda})^{2} \lambda I)=\frac{1}{\sqrt{\lambda}}\mathcal{N}(\vec{0}, R(\theta)^{2}\sigma^{2}I)$ random variable. Thus, the difference of these gradients converges to $\vec{0}$ as $\lambda\rightarrow\infty$ by the law of large numbers.

\section{The Distribution of $||\alpha - \theta|| = ||\mathcal{N}(\vec{0},\sigma^{2} I)||$\label{distributionof||}}
In Section \ref{proof}, we assumed that $||\alpha - \theta||$ was approximately constant. In this section, we validate that assumption.\\
The distribution of $||\alpha- \theta|| = ||\mathcal{N}(\vec{0},\sigma^{2} I)||$ is the $\sigma\chi$ distribution\footnote{$\sigma$ times the $\chi$ distribution}, which has mean
\begin{align}
\mu = \sigma\frac{\sqrt{2}\Gamma(\frac{n+1}{2})}{\Gamma(\frac{n}{2})},\label{mean}
\end{align}
and variance
\begin{align}
s^{2} = 2\sigma^{2}\left(\frac{\Gamma(\frac{n+2}{2})}{\Gamma(\frac{n}{2})} - \left(\frac{\Gamma(\frac{n+1}{2})}{\Gamma(\frac{n}{2})}\right)^{2}\right).
\end{align}
Consider the identities\cite{tricomi1951}
\begin{align}
\frac{\Gamma(z+a)}{\Gamma(z+b)} &= z^{a-b}\left[1 + \frac{(a-b)(a + b - 1)}{2z} + O(|z|^{-2})\right]\label{gammaab}\\
\frac{\Gamma(z+1)}{\Gamma(z)} &= z
\end{align}
And now, let us investigate the limits of $\mu$ and $s^{2}$ as $n\rightarrow\infty$.

\subsection{$\lim_{n\rightarrow\infty}\mu$}
Using \ref{gammaab}, we find
\begin{align}
	\mu &= \sigma\sqrt{2}\left((\frac{n}{2})^{\frac{1}{2}}\left[1 + \frac{(a-b)(a+b-1)}{n} + O(|\frac{n}{2}|^{-2})\right]\right)\\
	&=\sigma\sqrt{2}\left(\sqrt{\frac{n}{2}} + O(\sqrt{\frac{1}{n}})\right)\\
	&\implies\lim_{n\rightarrow\infty}\mu = \lim_{n\rightarrow\infty}\sqrt{n\sigma^{2}}
\end{align}
that is, $\mu\rightarrow\infty$ as $n\rightarrow\infty$.


\subsection{$\lim_{n\rightarrow\infty}s^{2}$}
\begin{align}
	s^{2} &= 2\sigma^{2}\left(\frac{\Gamma(\frac{n+2}{2})}{\Gamma(\frac{n}{2})} - \left(\frac{\Gamma(\frac{n+1}{2})}{\Gamma(\frac{n}{2})}\right)^{2}\right)\\
	&= 2\sigma^{2} \left(\frac{n}{2} - \left((\frac{n}{2})^{\frac{1}{2}}\left[1 + \frac{(\frac{1}{2})(\frac{-1}{2})}{n} + O(|\frac{n}{2}|^{-2})\right]\right)^{2}\right)\\
	&= 2\sigma^{2} \left(\frac{n}{2} - \left((\frac{n}{2})^{\frac{1}{2}} + \frac{-(\frac{n}{2})^{\frac{1}{2}}}{4n} + (\frac{n}{2})^{\frac{1}{2}}O(|\frac{n}{2}|^{-2})\right)^{2}\right)\\	
	&= 2\sigma^{2} \left(\frac{n}{2} -
	\left(\frac{n}{2} + 
	\frac{-n}{8n^{2}} +
	O(|\frac{n}{2}|^{-3}) -
	\frac{1}{4} +
	2\frac{n}{2}O(|\frac{n}{2}|^{-2}) -
	\frac{1}{4}O(|\frac{n}{2}|^{-2})
	\right)\right)\\
	&= 2\sigma^{2} \left(\frac{n}{2} - \left(\frac{n}{2} - \frac{1}{4} + \frac{-1}{8n} + 2O(n^{-1}) - \frac{1}{4}O(|\frac{n}{2}|^{-2}) + O(|\frac{n}{2}|^{-3})\right)\right)\\
	&= 2\sigma^{2} \left(\frac{n}{2} - \left(\frac{n}{2} - \frac{1}{4} + O(\frac{1}{n})\right)\right)\\
	&= 2\sigma^{2} \left(\frac{1}{4} - O(\frac{1}{n})\right)\\
	&\implies \lim_{n\rightarrow\infty}s^{2} = \frac{\sigma^{2}}{2}
\end{align}

\subsection{$\lim_{n\rightarrow\infty}\frac{s^{2}}{\mu}$}
Considering the two limits determined above,
\begin{align}
	&\lim_{n\rightarrow\infty}\frac{s^{2}}{\mu} = 0\\
	\iff &\lim_{n\rightarrow\infty}\int_{0}^{\infty}\frac{(x-\mu)^{2}}{\mu}f_{\chi(n)}dx = 0
\end{align}
and thus the $\frac{\sigma \chi(n)}{\mu}$ distribution converges in measure to the Dirac Delta Function. This satisfies the requirement of Section \ref{proof} of being (or becoming) ``closely distributed''. As the dimension of the vector under optimization in Section \ref{proof} increases, the approximation becomes \textit{almost surely} good.
\section{Closing Remarks}

\subsection{The Shape of $\mathcal{N}(0, I)$ in $n$ Dimensions}
It should be noted that the proof in Section \ref{distributionof||} shows that $\mathcal{N}(0, I)$ approaches the uniform distribution over the surface of an $n$-sphere, in the sense that the norm of $\mathcal{N}(0, I)$ approaches a constant, and the angular component is uniformly distributed. This provides some insight as to the nature of the optimization of this distribution; the ``cloud'' being optimized by the distribution-gradient of Evolution Strategies has a well-defined shape. In this sense, this paper resolves some of the conceptual ambiguity of ``cloud'' optimization. If the object under optimization is the surface of a sphere, it is intuitive that if the policy is continuous in its parameterization, and the sphere is small, that optimization of the cloud should be identical to optimizing the center of that sphere.

Further, because this structure for the noise only becomes clear in high-dimension contexts, this, along with choice of $\sigma$, explains the results of visualizations of ES in 2 dimensions used in \textit{Evolution Strategies is more than Just a Traditional Finite Differences Approximator}\cite{lehman2018}.

\subsection{\textit{Smallness}, Evolution Strategies, and $\nabla R(\theta)$\label{smallness}}
The size of a single perturbation in Evolution Strategies is given by \ref{mean}, and is much larger than $\sigma$ in general. Whether this is small in the context of the differentiability of $R$ is of interest; if it is, then Evolution Strategies approximates the policy gradient in addition to the distribution-gradient (i.e. the gradients are identical). If it is not, then it approximates only the distribution-gradient. Put another way, the gradients of ES and FD are always similar in the limit, but their shared gradient only approximates the policy gradient if $\sigma$ is small.

\section{Conclusion}
Evolution Strategies and Finite Differences converge as the dimension of an optimization problem increases. For high-dimension problems, such as those in deep learning, the algorithms are approximately equivalent. If $\sigma$ is small, then the shared gradient of ES and FD approximates the policy gradient. When $\sigma$ is not small, the gradient still formally follows the ``distribution gradient''. Given the effectiveness of the policy gradient, and the convergence of ES to the policy gradient for small $\sigma$, it is possible that the strength of ES observed by Salimans et al.\cite{salimans2017} \textit{is not due to the optimization of a distribution}, but to the long-used standard gradient (or ``policy gradient'', in the machine learning context).

It is also possible that the motivations (policy- and distribution- gradients) are effective across different ranges of $\sigma$, although there is no evidence for this in the literature. Such proof is only possible on a problem-by-problem basis, and would require proving 1) that at a given $\sigma$ the ES-gradient \textit{does not} approximate the true gradient, and 2) that at this value of $\sigma$ optimization is still possible using the ES-gradient. In light of the fact that this would imply that Finite Differences would be equally effective at these (non-small) values of $\sigma$, the authors consider this possibility unlikely.

This result raises doubts as to the advance that Evolution Strategies was thought to represent for non-policy gradient algorithms as optimizers in Reinforcement Learning. If the success of Evolution Strategies is attributable to its similarity to Finite Differences, then Evolution Strategies represents only a step forward for a certain method of direct approximation of policy gradients in the field of Reinforcement Learning.

\printbibliography
\end{document}